\documentclass[letterpaper, 10 pt, conference]{ieeeconf}  % Comment this line out if you need a4paper

\IEEEoverridecommandlockouts                              % This command is only needed if 
\usepackage[T1]{fontenc}        % Critical for font encoding
\usepackage{textcomp}           % Additional text symbols
\usepackage{mathptmx}           % Times font for text and math

\usepackage{graphics} % for pdf, bitmapped graphics files
\usepackage{epsfig} % for postscript graphics files
\usepackage{mathptmx} % assumes new font selection scheme installed
\usepackage{amsmath} % assumes amsmath package installed
\usepackage{amssymb}  % assumes amsmath package installed
\usepackage{graphicx}
\usepackage{multicol} % Add this to your preamble
\usepackage{xcolor}
\usepackage{makecell}
\usepackage{multirow}
\usepackage{hyperref} 
\usepackage{newtxmath} % 支持 Times 字体的数学符号
\usepackage{pifont}
\definecolor{green}{rgb}{0, 1, 0}
\definecolor{red}{rgb}{1, 0, 0}
\usepackage{graphicx}
\usepackage{array}
\usepackage{rotating}
\usepackage{pifont}
\usepackage[table]{xcolor}
\usepackage{tabularx}
\usepackage{pifont}
\usepackage{xcolor}
\usepackage{booktabs}
\usepackage{array}
\usepackage[table]{xcolor}
\usepackage{tabularx}
\usepackage{pifont} % for \ding{51} and \ding{55}
\newcommand{\cmark}{\textcolor{green!70!black}{\ding{51}}}
\newcommand{\xmark}{\textcolor{red}{\ding{55}}}

\title{\LARGE \bf
MLLM-Fabric: Multimodal Large Language Model-Driven Robotic Framework for Fabric Sorting and Selection}

\author{Liman Wang$^{1}$, Hanyang Zhong$^{1}$, Tianyuan Wang$^{1}$, Shan Luo$^{2}$, and Jihong Zhu$^{1}$% <-this % stops a space
\thanks{*This work was supported by the China Scholarships Council Scholarship - University of York Joint Programme. (Corresponding author: Jihong Zhu)}% <-this % stops a space
\thanks{$^{1}$ Authors are with the School of Physics, Engineering and Technology, University of York, York YO10 5DD, United Kingdom. (Email addresses: liman.wang@york.ac.uk,  hanyang.zhong@york.ac.uk, tianyuan.wang@york.ac.uk, jihong.zhu@york.ac.uk)}%
\thanks{$^{2}$ Shan Luo is with Robot Perception Lab at the Department of Engineering, King’s College London, London WC2R 2LS, United Kingdom. (Email address: shan.luo@kcl.ac.uk)}%
}

\begin{document}

\maketitle
\thispagestyle{empty}
\pagestyle{empty}

%%%%%%%%%%%%%%%%%%%%%%%%%%%%%%%%%%%%%%%%%%%%%%%%%%%%%%%%%%%%%%%%%%%%%%%%%%%%%%%%
\begin{abstract}
Choosing appropriate fabrics is critical for meeting functional and quality demands in robotic textile manufacturing, apparel production, and smart retail. We propose MLLM-Fabric, a robotic framework leveraging multimodal large language models (MLLMs) for fabric sorting and selection. Built on a multimodal robotic platform, the system is trained through supervised fine-tuning and explanation-guided distillation to rank fabric properties.
We also release a dataset of 220 diverse fabrics, each with RGB images and synchronized visuotactile and pressure data. Experiments show that our Fabric-Llama-90B consistently outperforms pretrained vision-language baselines in both attribute ranking and selection reliability. Code and dataset are publicly available at \url{https://github.com/limanwang/MLLM-Fabric}.
\end{abstract}

%%%%%%%%%%%%%%%%%%%%%%%%%%%%%%%%%%%%%%%%%%%%%%%%%%%%%%%%%%%%%%%%%%%%%%%%%%%%%%%%
\section{Introduction}
Robotic fabric manipulation has attracted increasing attention in tasks such as material identification, physical property estimation, and deformable object handling~\cite{ClothManipulation}. Although visual and tactile sensing have improved perceptual capabilities, conventional classification methods often fail to meet practical selection requirements. For instance, fabrics labeled as ``cotton'' may vary significantly in softness, elasticity, and thickness, which directly impact downstream tasks including folding, dressing, and tool interaction.
Effective fabric selection demands reasoning over continuous material properties and context-specific relevance. Visual-only methods are limited in capturing compliance and friction cues, while prior multimodal approaches~\cite{Yuan,activecloth,vitac,Multimodal_zero-shot} often depend on unimodal embeddings, lack supervision on physical attributes, or treat fabrics as static object categories, without enabling structured comparison or functional decision-making. These limitations highlight the need for a new paradigm in robotic material understanding.

\begin{figure}[t]
\centering
\includegraphics[width=1\linewidth]{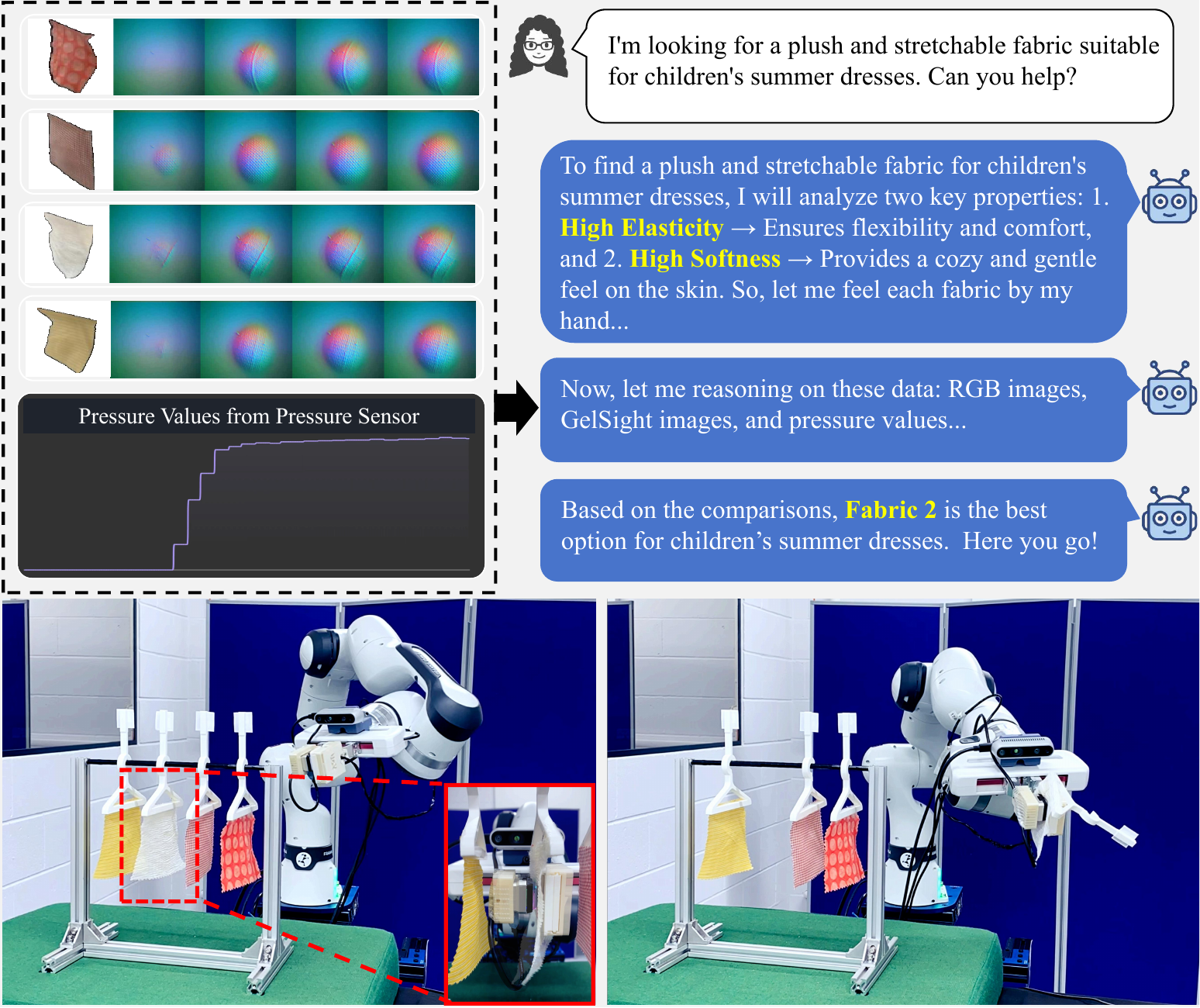}
\caption{The robotic system uses multimodal perception and reasoning to select fabrics based on user needs, choosing pure cotton seersucker (the white one on the hanger) as the optimal summer clothing. GelSight images are sequential frames with corresponding pressure sensor data. The figure shows a pressure trend graph for one fabric as an example.}
\label{fig:fig1}
\end{figure}

As shown in Fig.~\ref{fig:fig1}, this work introduces a physically grounded and task-aware robotic framework that formulates fabric selection as a property-specific pairwise comparison problem. Rather than assigning absolute class labels, the system learns to determine which of two materials better satisfies a specified property, such as softness or elasticity. This relative reasoning supports context-sensitive and function-oriented decisions, which are essential for real-world applications like adaptive grasping and clothing recommendation, where material behavior is more critical than category identity.
At the core of our system is a multimodal large language model trained on synchronized RGB, tactile, and pressure data. The model is optimized through pairwise preference supervision to enable interpretable, physically grounded material comparisons. Additionally, we apply explanation-guided knowledge distillation to enhance reasoning clarity and allow abstention in uncertain cases.
To support this formulation, we construct a dataset of 220 real-world fabric samples, each annotated with four physical attributes and paired with co-registered RGB images, GelSight sequences, and pressure readings. This dataset enables reproducible benchmarking of property-level material comparison under realistic sensory conditions.

\begin{figure*}[t]
\centering
\includegraphics[width=1\linewidth]{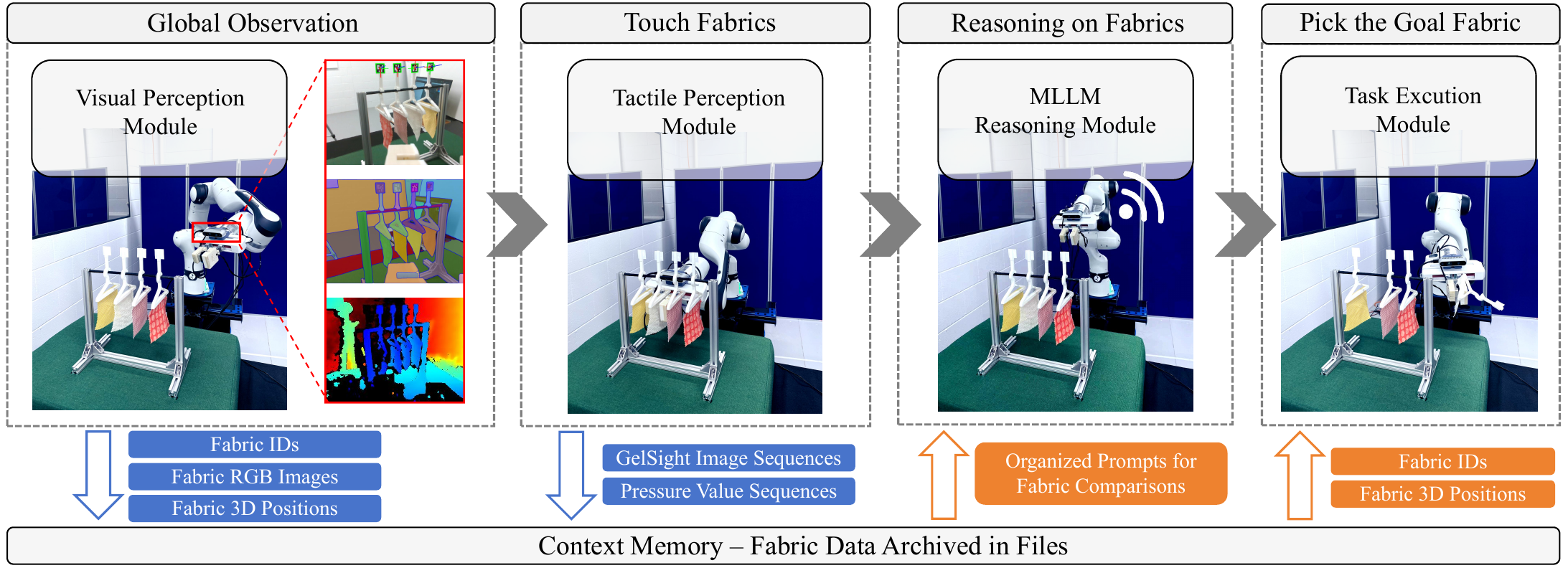}
\caption{Fabric selection robotic system workflow. The system follows a four-stage pipeline. First, a \textbf{Visual Perception Module} captures fabric IDs, RGB images, and 3D positions. Second, a \textbf{Tactile Perception Module} records GelSight image sequences and pressure values. Third, an \textbf{MLLM Reasoning Module} processes multimodal data to compare fabrics. Finally, a \textbf{Task Execution Module} selects and retrieves the goal fabric using stored IDs and positions. Fabric data is archived for efficient decision-making.}
\label{fig:fig2}
\end{figure*}

The main contributions of this work are as follows:
\begin{itemize}
    \item A new problem formulation that recasts robotic material selection as a property-specific pairwise comparison task, supporting functional reasoning under ambiguity and task variability.
    \item A multimodal large language model framework trained with pairwise preference supervision and explanation-guided knowledge distillation, enabling interpretable and abstention-aware predictions from tactile, visual, and force modalities.
    \item A robotic system that performs property-level reasoning for deformable object selection, where relative material behavior provides more informative guidance than discrete class identity.
    \item A real-world dataset of 220 fabrics with co-registered RGB, GelSight, and pressure data, annotated with semantic physical properties, to support future research in semantic perception and robotic material understanding.
\end{itemize}

\section{Related Work}
While prior work has explored visuotactile integration, most approaches treat fabric understanding as a classification or retrieval task. In contrast, our work frames it as property-specific, pairwise reasoning for task-conditioned selection, requiring different forms of supervision and inference.
Yuan et al.~\cite{Yuan} trained CNNs on RGB, depth, and GelSight data to learn modality-invariant embeddings for individual fabrics, enabling cross-modal inference but not supporting comparison or task-driven selection. Their subsequent work~\cite{activecloth} developed an autonomous robot system to classify fabric properties using tactile images collected via Kinect-guided squeezing, but still lacked multimodal fusion or pairwise reasoning.
ViTac~\cite{vitac} used deep maximum covariance analysis to embed visual and tactile data into a shared latent space for texture classification, but did not address functional or mechanical properties. Cao et al.~\cite{Multimodal_zero-shot} proposed a generative method to synthesize tactile features from vision and language, enabling zero-shot recognition, but only produced coarse semantic labels without supporting fine-grained attribute ranking or selection.

Recent work integrates tactile inputs into large-scale vision-language frameworks. TVL~\cite{TVL} trained a touch-vision-language encoder on pseudo-labeled data for open-vocabulary material recognition. However, it lacks physical supervision, material comparison, or structured selection capabilities. UniTouch~\cite{UniTouch} aligned multiple tactile sensors with vision-language spaces, supporting cross-modal generalization, but omitted property supervision and any form of reasoning or ranking. Among existing methods, Octopi~\cite{Octopi} is the most closely related to our work. It introduces visuotactile prompting for property inference, but operates at the object level by predicting coarse attributes of entire items. In contrast, our framework performs material-level reasoning over localized fabric properties and supports task-conditioned comparison and selection.

Table~\ref{tab:tab1} summarizes eight representative methods across six functional axes relevant to material selection. Our method is the only one to unify all six capabilities in a fabric-specialized system. These axes span from task scope (fabric specificity), to perceptual coverage (vision-touch fusion), to reasoning ability (property comparative inference and ranking), and finally to task conditioning and language-based generalization. By framing material understanding as a grounded, pairwise reasoning task enriched with explanation generation, our system bridges the gap between semantic perception and task-driven robotic manipulation, enabling interpretable and goal-aligned material decisions.

\begin{table}[t]
\caption{
Comparison of representative methods across capabilities: fabric-specific focus (Fabric), multimodal fusion (MM), comparative reasoning (CR), ranking (RK), task awareness (TA), and use of multimodal language models (MLLM). \textsuperscript{†}Refers to the study titled Active Clothing}
\label{tab:tab1}
\centering
\setlength{\tabcolsep}{7pt}
\renewcommand{\arraystretch}{1.1}
\begin{tabular}{lcccccc}
\hline
\textbf{Method} & Fabric & MM & CR & RK & TA & MLLM \\
\hline
Yuan et al.~\cite{Yuan} & {\ding{51}} & {\ding{51}} & {\ding{55}} & {\ding{55}} & {\ding{55}} & {\ding{55}} \\
Yuan et al.\textsuperscript{†}~\cite{activecloth} & {\ding{51}} & {\ding{55}} & {\ding{55}} & {\ding{55}} & {\ding{55}} & {\ding{55}} \\
ViTac~\cite{vitac} & {\ding{51}} & {\ding{51}} & {\ding{55}} & {\ding{55}} & {\ding{55}} & {\ding{55}} \\
Cao et al.~\cite{Multimodal_zero-shot} & {\ding{51}} & {\ding{51}} & {\ding{55}} & {\ding{55}} & {\ding{55}} & {\ding{55}} \\
TVL~\cite{TVL} & {\ding{55}} & {\ding{51}} & {\ding{55}} & {\ding{55}} & {\ding{51}} & {\ding{51}} \\
UniTouch~\cite{UniTouch} & {\ding{55}} & {\ding{51}} & {\ding{55}} & {\ding{55}} & {\ding{51}} & {\ding{51}} \\
Octopi~\cite{Octopi} & {\ding{55}} & {\ding{51}} & {\ding{55}} & {\ding{55}} & {\ding{51}} & {\ding{51}} \\
\textbf{Ours} & {\ding{51}} & {\ding{51}} & {\ding{51}} & {\ding{51}} & {\ding{51}} & {\ding{51}} \\
\hline
\end{tabular}
\end{table}

\section{Methodology}

\subsection{Visual Data Processing and Fabric Localization}
The overall workflow is illustrated in Fig.~\ref{fig:fig2}. An Intel RealSense D435 camera mounted on the robotic arm captures RGB and depth data to enable fabric localization and segmentation. Each fabric sample is labeled with an ArUco tag, providing a unique ID for pose tracking on the hanger. Computer vision techniques are used to estimate the 3D position and orientation of each sample. Depth data along the Z-axis is fused with ArUco-based tracking to improve pose estimation under shape variation and hanger misalignment, ensuring robust visual localization for downstream multimodal processing.
After localization, the Segment Anything Model (SAM)~\cite{Segment_Anything} performs pixel-level segmentation to isolate fabric samples from the background and neighboring objects. Each segmented image is resized to 224$\times$224 pixels and stored along with metadata tuples \( (F_i, P_i, I_i) \), where \( F_i \) is the fabric ID, \( P_i \in \mathbb{R}^3 \) is the 3D position vector, and \( I_i \) is the segmented RGB image. This structured representation maintains modality alignment and supports subsequent multimodal reasoning and property prediction.

\begin{figure}[t]
\centering
\includegraphics[width=1\linewidth]{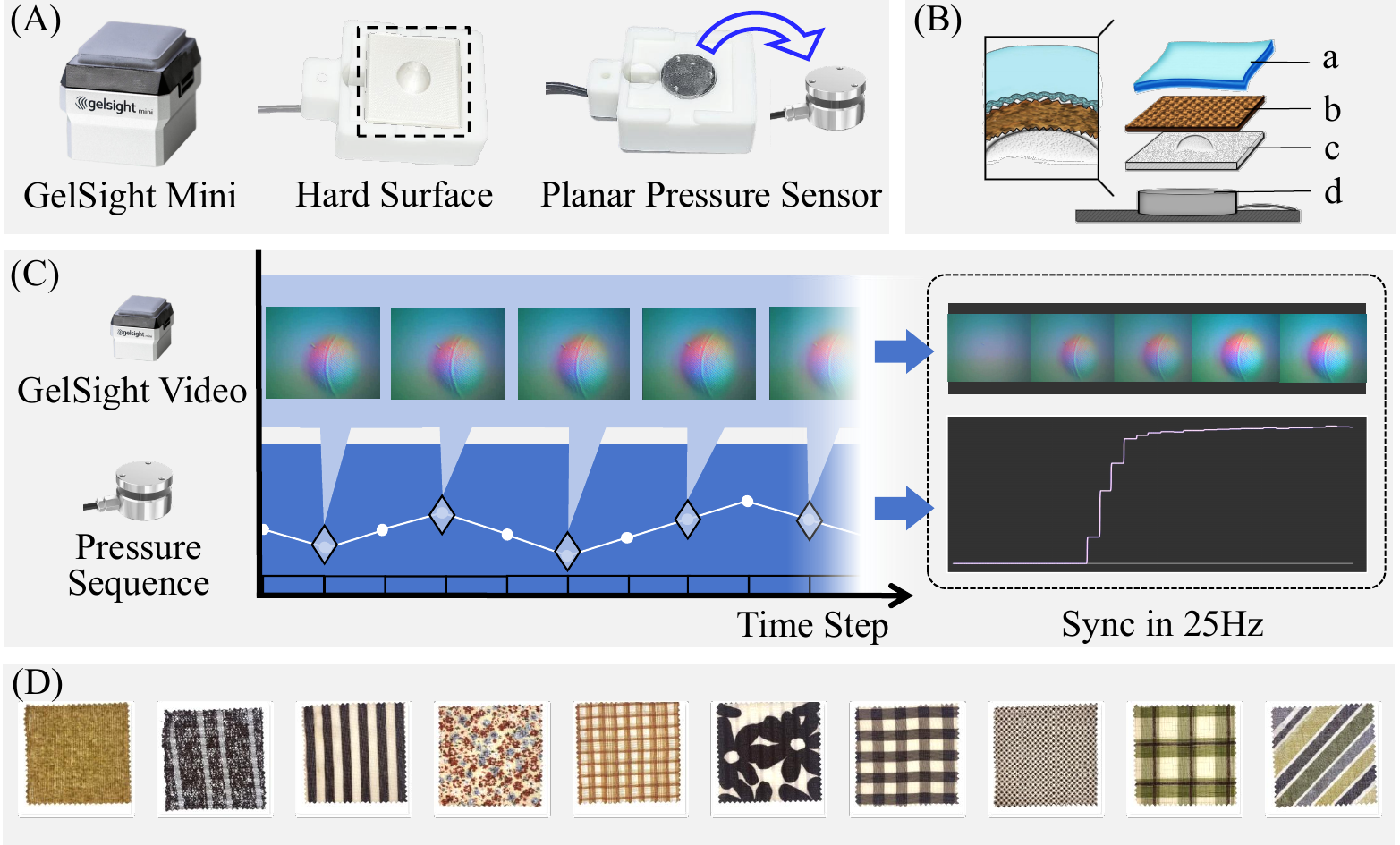}
\caption{(A) Overview of the tactile sensing setup, including the GelSight Mini sensor and a planar pressure sensor mounted on a hard surface. (B) The layered structure during pressing: (a) gel layer, (b) fabric sample, (c) hard surface, and (d) pressure sensor. (C) Synchronization of tactile image sequences from the GelSight sensor with corresponding pressure measurements at a frequency of 25 Hz. (D) A selection of ten representative fabric samples from a dataset containing 220 unique fabrics.}
\label{fig:fig3}
\end{figure}

\subsection{Fabric Tactile Sensing and Force-labeled Tactile Data}
A dual-sensor setup is used for fabric tactile sensing (Fig.~\ref{fig:fig3}A): the left fingertip of the Franka Hand mounts a GelSight Mini~\cite{gelsight} for high-resolution visuotactile imaging, while the right fingertip holds a strain-based planar pressure sensor with a dome surface for direct force measurement (Fig.~\ref{fig:fig3}B).
Although the GelSight sensor can estimate force from surface deformation, its accuracy degrades significantly on fabrics due to sensitivity to local thickness, compliance~\cite{activecloth}, and device-level variation~\cite{gelsight}. To avoid such uncertainty, we use a strain-gauge pressure sensor (HZC-30B, Chengying Sensor, China) mounted beneath a rigid planar surface. It has a 0--10~kg range, 1.0--1.5~mV/V sensitivity, $\pm$0.3\% F.S. nonlinearity, $\pm$0.5\% F.S. creep, and operates from $-30\,^\circ\mathrm{C}$ to $70\,^\circ\mathrm{C}$ with 200\%~F.S. overload tolerance.
The two sensors are mounted on opposing fingertips to compress the fabric simultaneously. By Newton’s third law, the pressure reading equals the normal force on the GelSight surface. This setup enables synchronized recording of force and deformation without relying on image-based force estimation. The resulting stream forms \textbf{force-labeled tactile data}, where each GelSight frame is aligned with a corresponding pressure value. We refer to these pressure readings as “force” throughout. This alignment replaces the need for GelSight force calibration and provides accurate ground-truth supervision during training.

\subsection{Multimodal Data Collection and Fusion Strategy}
We collect data from 220 fabrics using a Franka arm, gripper, RealSense D435, GelSight Mini, and a planar pressure sensor. Each fabric is pressed twice for 10 seconds, with moderate-speed approach and slow closure to ensure stable contact. GelSight and pressure data are sampled at 25~Hz and time-synchronized (Fig.~\ref{fig:fig3}C), pairing each frame with a force reading. RGB examples in Fig.~\ref{fig:fig3}D illustrate visual diversity. The dataset includes RGB images, GelSight sequences, and aligned pressure values. GelSight captures spatial deformation related to texture and compliance, while the pressure sensor provides scalar force measurements. Fusion preserves both deformation context and contact intensity.
To integrate this heterogeneous data, we extract feature embeddings from each modality. RGB and GelSight images are processed by a frozen visual encoder, and pressure signals are converted into language-compatible embeddings via the model’s input projection layers. The resulting embeddings are unified in a shared token space and jointly processed by a multimodal transformer. This early-fusion strategy enables learning of cross-modal cues during training and inference, following recent vision-language models that align diverse modalities through token-level reasoning~\cite{liu2023llava}.
Each fabric is also annotated with categorical labels for elasticity, softness, thickness, and texture by human experts. These annotations support supervised fine-tuning and evaluation.

\subsection{Supervised Fine-Tuning}

In our study, supervised fine-tuning of an MLLM using directly labelled training data is referred to as Directly Supervised Fine-Tuning (D-SFT). Unlike knowledge distillation, which refines predictions based on teacher model outputs, D-SFT directly learns from human-annotated comparisons, ensuring explicit alignment of model parameters with expert judgments.

\subsubsection{Supervised Fine-Tuning Objective}

The goal of supervised fine-tuning is to train the model to rank fabric properties based on pairwise comparisons. Since each fabric sample is unique, no two fabrics share identical attributes. The model's primary task is to establish a ranking, with cases of minimal or indistinguishable differences classified as inconclusive.

The training objective is formulated as a classification task, where the model outputs a probability distribution over the three possible outcomes. The learning process is guided by a cross-entropy loss function:
\begin{equation}
L_{\text{train}} = \sum_{i=1}^{N} \sum_{q \in Q'} L_{\text{CE}} \left( g(z_i, \phi), y_{i,q} \right)
\end{equation}
where \( g(z_i, \phi) \) denotes the predicted probability distribution for fabric property \( q \), and \( y_{i,q} \) represents the corresponding ground truth label. The input \( z \) includes multimodal information, incorporating integrating visual, tactile, and pressure data to capture relevant fabric characteristics.

\subsubsection{LoRA-Based Fine-Tuning}
To enable efficient and task-specific adaptation, we adopt Low-Rank Adaptation (LoRA)~\cite{lora} for fine-tuning. LoRA introduces trainable low-rank matrices into the projection layers of the pre-trained language model, allowing effective parameter updates with minimal overhead. Formally, the adapted weights are defined as:
\begin{equation}
W' = W + \Delta W, \quad \Delta W = AB,
\end{equation}
where \( A \in \mathbb{R}^{d \times r} \) and \( B \in \mathbb{R}^{r \times d} \), with rank \( r \ll d \).
We explore two strategies for integrating LoRA into our framework: (1) applying fine-tuning to both the vision encoder and language module, and (2) applying LoRA-based fine-tuning solely to the language module while keeping the visual backbone frozen. The latter is motivated by findings in recent multimodal literature suggesting that pretrained visual encoders can retain strong generalization even when frozen, particularly in tasks involving limited domain-specific data~\cite{liu2023llava}.

\subsubsection{Inference Process}

During inference, the model processes a given pair of fabric samples and predicts the most probable ranking for the specified property. The decision is obtained by selecting the class with the highest predicted probability:
\begin{equation}
\hat{y}_q = \arg\max g(z, \phi)
\end{equation}
This pairwise prediction mechanism enables the model to differentiate fabric properties effectively based on direct comparisons.
D-SFT directly optimizes for pairwise classification using labelled data. While knowledge distillation benefits from leveraging additional synthetic training signals, D-SFT is constrained by the quality and availability of labelled comparisons. 

\begin{figure}[t]
\includegraphics[width=\linewidth]{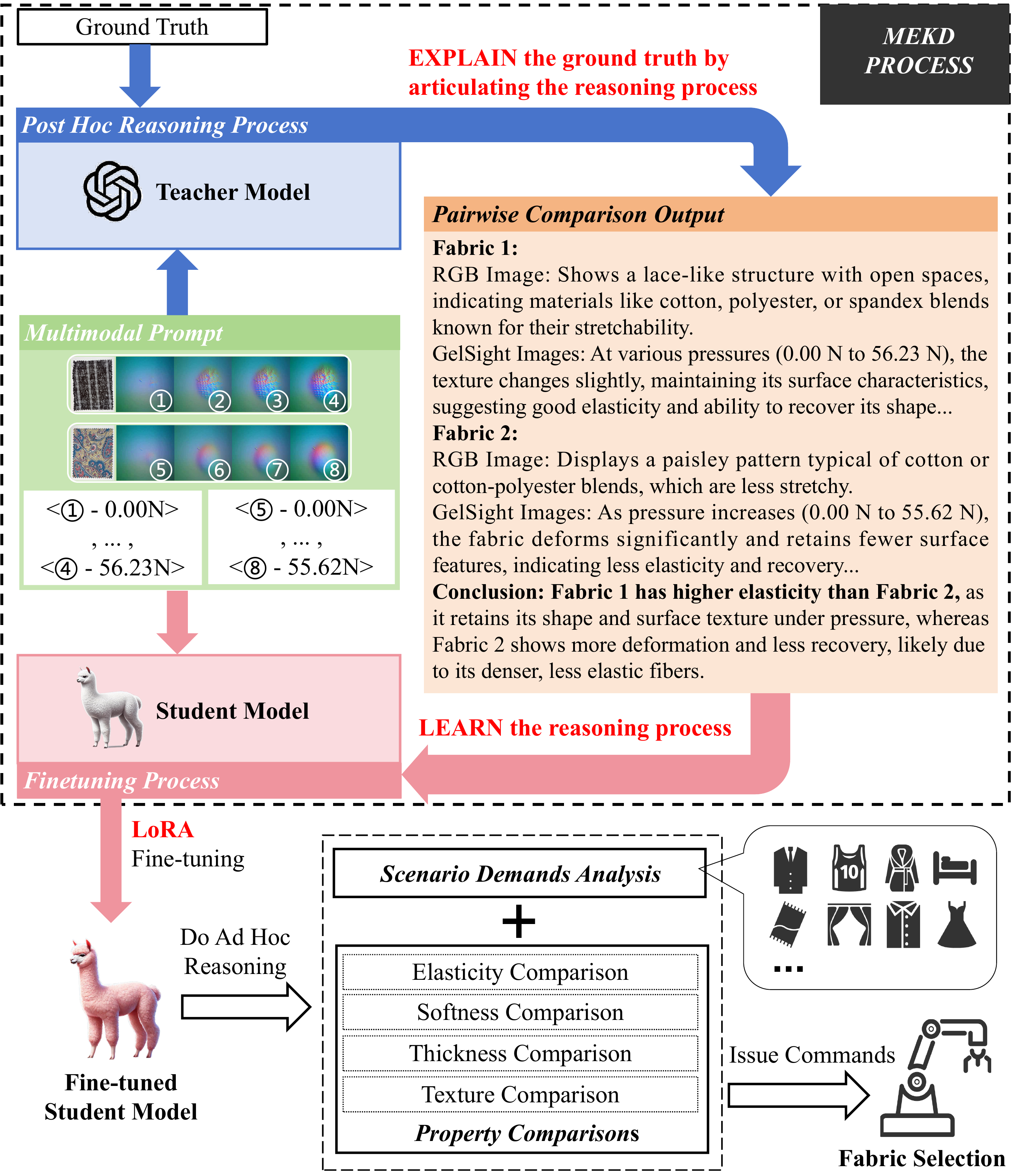}
\caption{\textbf{Multimodal Reasoning and Knowledge Distillation Pipeline.} The upper section illustrates Multimodal Explanation-Guided Knowledge Distillation, where a teacher model generates \textit{post hoc} explanations that are subsequently distilled into a student model. The lower section depicts the fabric property comparison and selection process, where the fine-tuned model performs \textit{ad hoc} reasoning to compare attributes and issue fabric selection commands.}
\label{fig:fig4}
\end{figure}

\subsection{Explanation-Guided Knowledge Distillation}
To enhance ranking accuracy and interpretability, we introduce a Multimodal Explanation-Guided Knowledge Distillation (MEKD) process that integrates expert-validated reasoning. GPT-4o~\cite{GPT-4}, being a closed-source model and significantly larger than Llama3.2-Vision-90B~\cite{llama}, is impractical for direct deployment due to its restricted accessibility and high computational costs. Instead, we distill GPT-4o's reasoning ability into the more efficient Llama3.2-Vision-90B, transferring its structured justifications through MEKD while preserving interpretability. As shown in Fig.~\ref{fig:fig4}, this process involves pairwise fabric comparisons and AI-generated rationales from GPT-4o, which are directly incorporated into the training pipeline to enhance model supervision and improve the reliability of the distilled model.

\subsubsection{Post Hoc Reasoning}
The teacher model first generates \textit{post hoc} ranking explanation content based on ground truth data. Human experts refine these explanations to improve clarity and reliability before using them to fine-tune the student model. This process is formulated as:
\begin{equation}
Q_{\text{post hoc}} = \text{MLLM}({z}, Q_{\text{inst}}, \tilde{r})
\end{equation}
where MLLM($z, Q^{\text{inst}}, r$) denotes the Multimodal Large Language Model module that generates post hoc ranking explanations based on the input pair $z$, the instruction $Q^{\text{inst}}$, and the expert-validated ranking $r$.
This approach enhances interpretability, aligns model predictions with expert knowledge, and provides structured supervision to improve generalization.
\subsubsection{Ad Hoc Reasoning}
In the absence of ground truth labels, the student model generates \textit{ad hoc} reasoning to approximate expert explanations. To enforce consistency, we minimize the discrepancy between \textit{post hoc} and \textit{ad hoc} outputs via a cross-entropy loss:
\begin{equation}
\hat{\phi} = \arg \min_{\phi} \mathcal{L}_{\text{CE}}(Q_{\text{ad hoc}}, Q_{\text{post hoc}})
\end{equation}
where \( Q_{\text{ad hoc}} \) represents the model’s self-generated reasoning content.
By aligning self-generated justifications with expert-reviewed rationales, the model improves reasoning consistency and robustness, enabling adaptation to unseen data.

\section{Experimental Setup}

\subsection{Fabric Selection Scenario Design and Evaluation}
We evaluate the models' ability to reason about fabric properties in ten expert-designed selection scenarios, covering diverse applications such as apparel, home textiles, and industrial fabrics. Each scenario is designed to assess the model’s ability to differentiate fabrics based on attributes like elasticity, softness, thickness, and texture, requiring both simple and multi-attribute reasoning.

We compare Fabric-Llama (Llama-Vision-90B fine-tuned with MEKD) against the original Llama-Vision-90B and GPT-4o. In these unseen scenario evaluations, models analyze multimodal data to recommend the most suitable fabric based on expert-defined criteria. Accuracy is measured by comparing their selections with expert judgments.
The scenarios are structured to reflect real-world selection challenges, with increasing complexity when properties are closely ranked or when multiple attributes must be considered simultaneously. This setup evaluates whether fine-tuning enhances the model’s ability to generalize and make precise selections beyond basic classification.

\subsection{Model Training and Evaluation Setup}

Each fabric is annotated with a four-dimensional attribute vector $\mathbf{a} = [a_1, a_2, a_3, a_4] \in \{0,1,2\}^4$, representing elasticity, softness, thickness, and texture. Attributes are discretized into three levels: 0 (low), 1 (moderate), and 2 (high). To improve supervision clarity, we retain only level-0 and level-2 samples for training, discarding moderate-level cases to avoid ambiguity and enhance contrast in pairwise learning.

We use two evaluation settings. In the seen-fabric setting, 200 fabrics are used to generate 24,000 extreme-level comparison pairs. From these, 4,000 pairs are sampled for training and 400 for validation, evenly distributed across attributes. Fabric samples may recur, but pair combinations are unique across sets. The unseen-fabric setting uses 20 held-out fabrics with the same protocol and ensures no overlap at the fabric level. To evaluate generalization beyond binary extremes, we also assess the model on fine-grained comparisons between adjacent levels (0 vs. 1 and 1 vs. 2), which are excluded from training.
To handle uncertainty in ambiguous cases~\cite{LLM_Abstention}, we include an abstain option in all prompts. Though not a focus of primary evaluation, we test it on 50 curated duplicate pairs. Our fine-tuned models abstained in fewer than 50\% of cases, while GPT-4o abstained in over 90\%, indicating more conservative behavior. This suggests current training lacks duplicate-aware supervision and motivates future improvements.

We evaluate Llama3.2-Vision-90B for its strong multimodal reasoning, and DeepSeek Janus-Pro-7B as a lightweight baseline. GPT-4o is excluded due to its closed-source nature and reliance on non-reproducible APIs. Fine-tuning is performed on high-performance hardware: Llama3.2-Vision-90B is trained on an NVIDIA A800 (80~GB RAM) for 6.5 hours; Janus-Pro-7B and Llama3.2-Vision-11B are trained on an NVIDIA A6000 (48~GB RAM) for 3 and 5 hours, respectively.

\subsection{Evaluation Metrics}
We use three metrics to evaluate model performance: Property-wise Accuracy, Average Accuracy, and Prediction Skewness.
\textbf{Property-wise Accuracy (ACC)} measures correctness for each fabric property and is defined as:
\begin{equation}
\text{ACC} = \frac{TP + TN}{TP + TN + FP + FN},
\end{equation}
where $TP$, $TN$, $FP$, and $FN$ denote true/false positives and negatives, respectively.
\textbf{Mean Accuracy (mean ACC)} is the mean ACC across the four fabric properties and reflects overall ranking performance.
To quantify prediction balance, we define \textbf{Prediction Skewness (SK)} as:
\begin{equation}
\text{SK} = 2 \times \left| p_0 - 0.5 \right| \times 100,
\end{equation}
where $p_0$ is the proportion of predictions assigned to category 0. A SK of 0 indicates perfectly balanced outputs; higher values indicate increasing bias toward one class. Since ground-truth labels are evenly distributed (50\% per class), lower SK is preferred as it reflects greater prediction balance.

\section{Experimental Results}

\begin{figure*}[t]
    \centering
    \includegraphics[width=0.75\linewidth]{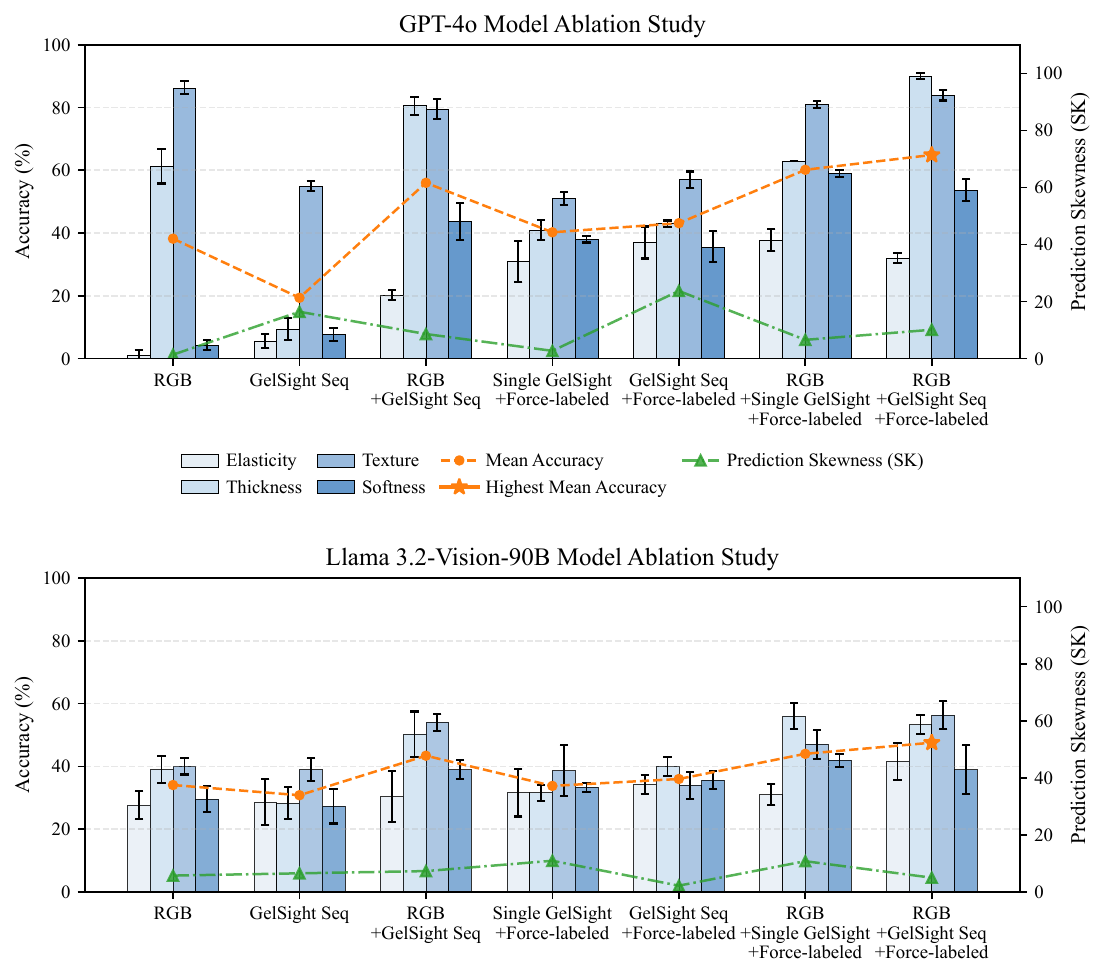}
    \caption{
    Ablation study of input modalities for pairwise property comparison using GPT-4o (top) and Llama3.2-Vision-90B (bottom). Bars show attribute-wise accuracy (\%), dashed lines indicate mean accuracy, and dotted lines show prediction skewness (SK; lower is better). Error bars denote standard error over three runs. “Single GelSight” = single-frame GelSight; “GelSight Seq” = Gelsight sequential frames.
    }
    \label{fig:fig5}
\end{figure*}
% \usepackage{multirow}
% \usepackage{graphicx}
\begin{table*}[]
\centering
\fontsize{5}{6}\selectfont
\resizebox{\textwidth}{!}{%
\begin{tabular}{llcccccc}
\hline
\textbf{Model} & \textbf{Fine-tuning} &
  \textbf{Elasticity ACC} $\uparrow$ &
  \textbf{Thickness ACC} $\uparrow$ &
  \textbf{Texture ACC} $\uparrow$ &
  \textbf{Softness ACC} $\uparrow$ &
  \textbf{mean ACC} $\uparrow$  &  \textbf{SK $\downarrow$}  \\ \hline
UniTouch-7B & None &
  0 & 0 & 0 & 0 & 0 & --\\
Octopi-13B & None &
  23 & 0 & 2 & 0 & 6 & 34.5\\
Octopi-13B & D-SFT &
 2 & 3 & 21 & 5 & 7.75 & 60.9\\
Octopi-13B & MEKD &
  4 & 5 & 40 & 0 & 12.25 & 78.7\\
DeepSeek Janus-Pro-7B & None &
  28 & 46 & 40 & 41 & 39 & 84.4\\
DeepSeek Janus-Pro-7B & D-SFT &
  50 & 50 & 50 & 50 & 50 & 100\\
Llama3.2-Vision-11B & None &
  53 & 46 & 44 & 53 & 49 & 60\\
Llama3.2-Vision-11B & D-SFT &
  50 & 51 & 45 & 42 & 47 & 31.4\\
Llama3.2-Vision-90B & None &
  46 & 56 & 52 & 35 & 47 & 2.1\\
Llama3.2-Vision-90B & D-SFT &
  \textbf{95} & \textbf{100} & 95 & 82 & 93 & 1\\ \hline
Fabric-Janus-7B & MEKD &
  50 & 50 & 50 & 47 & 47 & 5.5\\
Fabric-Llama-11B & MEKD &
  51 & 59 & 52 & 50 & 53 & 33.1\\
Fabric-Llama-90B* & MEKD (full fine-tuning)	 &
  55 & 93 & 81 & 62 & 73 & 4.5\\
  \textbf{Fabric-Llama-90B} & MEKD (frozen vision encoder) &
  90 & \textbf{100} & \textbf{97} & \textbf{89} & \textbf{94} & \textbf{0}\\ \hline
\end{tabular}%
}
\caption{Attribute-wise evaluation of different models trained on RGB, GelSight sequential frames, and synchronized force input for pairwise fabric property comparison. The table reports agreement accuracy with expert judgments on elasticity, thickness, texture, and softness, as well as mean accuracy and prediction skewness (SK). \textit{Asterisk (*) indicates full fine-tuning of both vision and language modules; models without * use MEKD with the vision encoder frozen.}}
\label{tab:tab2}
\end{table*}

\begin{table*}[t]
\centering
\fontsize{7}{9}\selectfont
\setlength{\tabcolsep}{2.0mm}{
\begin{tabular}{l|c|c|c|c|c|c}
\hline
\multirow{2}{*}{\textbf{Model}} 
& \multicolumn{2}{c|}{\textbf{Original Test}} 
& \multicolumn{2}{c|}{\textbf{Fine-grained Test}} 
& \multicolumn{2}{c}{\textbf{Unseen Fabrics}} 
\\ \cline{2-7}
& mean ACC $\uparrow$ & SK $\downarrow$ 
& mean ACC $\uparrow$ & SK $\downarrow$ 
& mean ACC $\uparrow$ & SK $\downarrow$ 
\\ \hline
Llama3.2-Vision-90B     
& 47 & 5.5 
& 39.7 & \textbf{12.8} 
& 42.8 & 13.6
\\
Llama3.2-Vision-90B + D-SFT     
& 93 & 1 
& 51.7 & 55.5 
& 55.7 & 7.4
\\
\textbf{Fabric-Llama-90B}    
& \textbf{94} & \textbf{0} 
& \textbf{52.2} & 54.5 
& \textbf{65.7} & \textbf{1.8}
\\ \hline
\end{tabular}}
\caption{
Generalization performance of selected models on three evaluation settings: (1) original test pairs from seen fabrics, (2) fine-grained comparisons with subtle attribute differences (level 1 vs.\ 0 or 2), and (3) comparisons on unseen fabrics. Metrics include agreement accuracy with expert pairwise rankings and prediction skewness (SK). All models use RGB, GelSight sequential frames, and force-labeled inputs.
}
\label{tab:tab3}
\end{table*}

\begin{table*}[t]
\scriptsize
\centering
\renewcommand{\arraystretch}{1.2}
\setlength{\tabcolsep}{3pt}

\begin{tabular}{p{9cm}cccccccccccc}
\toprule
\textbf{Scenario \& Properties} & \multicolumn{2}{c}{\textbf{Fabric-L90B}} & \multicolumn{2}{c}{\textbf{GPT-4o}} & \multicolumn{2}{c}{\textbf{LV90B}} & \multicolumn{2}{c}{\textbf{CNN IT}} & \multicolumn{2}{c}{\textbf{CNN IT*}} & \multicolumn{2}{c}{\textbf{Octopi-13B}} \\
 & Sort & Sel & Sort & Sel & Sort & Sel & Sort & Sel & Sort & Sel & Sort & Sel \\
\midrule
\rowcolor{gray!10}
Athletic wear for flexibility (Elasticity 2, Softness 2) & 100\% & \cmark & 33.3\% & \xmark & 50\% & \cmark & 50\% & \xmark & 50\% & \xmark & 16.7\% & \cmark \\
Summer dresses for children (Softness 2, Elasticity 2) & 83.3\% & \cmark & 20\% & \cmark & 0\% & \xmark & 62.5\% & \xmark & 62.5\% & \xmark & 16.7\% & \cmark \\
\rowcolor{gray!10}
Light draping fabric for home decor (Thickness 1, Softness 2) & 66.7\% & \cmark & 20\% & \xmark & 50\% & \xmark & 37.5\% & \xmark & 37.5\% & \xmark & 33.3\% & \xmark \\
Durable fabric for frequent-use furniture (Texture 2, Thickness 1) & 62.5\% & \cmark & 28.5\% & \xmark & 62.5\% & \cmark & 62.5\% & \cmark & 62.5\% & \cmark & 37.5\% & \cmark \\
\rowcolor{gray!10}
Protective fabric for work environments (Thickness 1, Elasticity 0) & 50\% & \cmark & 50\% & \xmark & 50\% & \xmark & 25\% & \xmark & 25\% & \xmark & 30\% & \xmark \\
Heavy-duty material for robust applications (Texture 2, Thickness 2) & 75\% & \cmark & 33.3\% & \xmark & 25\% & \xmark & 37.5\% & \cmark & 37.5\% & \cmark & 25\% & \cmark \\
\rowcolor{gray!10}
Soft lining fabric for formal suits (Thickness 0, Softness 2, Texture 0) & 80\% & \xmark & 42.8\% & \xmark & 30\% & \xmark & 50\% & \cmark & 50\% & \cmark & 10\% & \xmark \\
Lightweight smooth fabric for formal shirts (Thickness 0, Softness 2, Texture 0) & 90\% & \cmark & 45\% & \xmark & 9.1\% & \xmark & 75\% & \cmark & 75\% & \cmark & 36.3\% & \xmark \\
\rowcolor{gray!10}
Ultra-soft lightweight fabric for accessories (Softness 2, Thickness 0) & 50\% & \cmark & 50\% & \xmark & 16.7\% & \xmark & 75\% & \cmark & 75\% & \cmark & 16.7\% & \xmark \\
Thick, luxurious fabric for premium interiors (Thickness 2, Elasticity 0, Texture 2) & 61.5\% & \xmark & 23\% & \xmark & 15.3\% & \xmark & 41.6\% & \xmark & 41.6\% & \xmark & 38.4\% & \xmark \\
\bottomrule
\end{tabular}

\caption{
\textbf{Fabric selection performance across ten real-world scenarios.} 
\textbf{Sort} = property-specific pairwise ranking accuracy; 
\textbf{Sel} = final selection correctness. 
\textbf{Fabric-L90B} = proposed Fabric-Llama-90B model, 
\textbf{LV90B} = Llama-Vision-90B, 
\textbf{CNN IT / IT*} = interaction pipelines combining a CNN-LSTM tactile predictor with GPT-4o (CNN IT) or Llama-90B (CNN IT*) for downstream selection.
Attribute levels are encoded as 0 (low), 1 (moderate), and 2 (high). 
\textcolor{green}{\ding{51}} = correct selection; \textcolor{red}{\ding{55}} = failure.
}
\label{tab:tab4}
\end{table*}

\subsection{Fabric Property Sorting Experiment}

\subsubsection{\textbf{Impact of Data Modalities}}  
We conduct a systematic ablation across three input modalities: RGB, GelSight (single-frame or 4-frame sequences), and synchronized force labels. All meaningful combinations are evaluated to quantify each modality’s contribution to pairwise fabric property comparison. As shown in Fig.~\ref{fig:fig5}, the best performance is achieved using the full combination of RGB, sequential GelSight, and force input, reaching 65.2\% accuracy for GPT-4o and 47.6\% for Llama3.2. The poorest results occur with unlabeled GelSight sequences, yielding 19.4\% for GPT-4o and 30.8\% for Llama3.2. These findings indicate that sequential GelSight frames capture richer tactile cues than single-frame inputs, and that explicit force supervision is more critical for reliable multimodal reasoning than merely increasing input diversity.

\subsubsection{\textbf{Impact of Training Methodologies}}
We compare two training paradigms: Directly Supervised Fine-Tuning (D-SFT), which optimizes model predictions based on expert-labeled pairwise comparisons, and Multimodal Explanation-Guided Knowledge Distillation (MEKD), which transfers structured rationales from a larger teacher model.
As shown in Table~\ref{tab:tab2}, D-SFT substantially improves the performance of Llama3.2-Vision-90B (mean ACC: 93\%, SK: 1), and MEKD further enhances both accuracy (94\%) and prediction balance (SK: 0), indicating stronger generalization and reduced bias. 
We also evaluate full fine-tuning of both the vision encoder and the language module. This configuration achieves a mean ACC of 73\% and a skewness of 4.5, indicating lower accuracy and poorer output balance compared to the frozen-encoder variant. While less effective under limited data, full fine-tuning may still be useful given larger datasets or stronger regularization.
For the smaller Janus-Pro-7B, D-SFT results in highly imbalanced outputs (SK: 100), while MEKD reduces skewness to 5.5, confirming its effectiveness in improving prediction stability even for compact architectures.

\subsubsection{\textbf{Impact of Model Scale}}
Llama3.2-Vision-90B consistently outperforms Llama3.2-Vision-11B, reaching a mean ACC of 93–94\% compared to sub-60\% for the smaller model. For \textit{Elasticity}, D-SFT yields 95\% mean ACC, indicating that larger models can learn robust patterns even without post hoc reasoning. This suggests diminishing returns for post hoc supervision as model scale increases.

\subsubsection{\textbf{Model Generalization}}
As discussed in Section IV.B, the model is trained using coarse-grained attribute comparisons, such as level 2 versus level 0, in order to support generalization to fine-grained distinctions, including level 2 versus level 1 and level 1 versus level 0.
However, as shown in Table~\ref{tab:tab3}, both the post hoc reasoning model and the D-SFT model perform unreliably on fine-grained comparisons, with mean ACC remaining near 50\% and SK index exceeding 50, indicating high output uncertainty. 

We further validate our model on a 20-sample unseen fabric split (completely excluded from training), demonstrating robust generalization by evaluating performance on 140 pair-wise comparison questions, each contrasting level-2 and level-0 samples.
In this setting, the Llama3.2 baseline performs at chance level (42.8\%), while the D-SFT model reaches 55.7\% mean ACC with high uncertainty (SK = 7.4). Our full model (Fabric-Llama-90B) improves to 65.7\% mean ACC with a reduced SK index of 1.8, indicating a moderate but meaningful generalization gain. These results suggest that while the model can extract coarse-level patterns from seen data, its generalization to novel fabrics and fine-grained distinctions remains limited. Future work should explore targeted learning strategies to handle ambiguous attribute levels and shifts in fabric identity distribution.

\subsection{Fabric Selection Experiment}
This experiment evaluates each model’s ability to select suitable fabrics based on learned property reasoning. Unlike isolated attribute ranking, this task tests real-world decision-making. Fabric-Llama-90B, fine-tuned for this task, achieved the highest agreement with expert choices, correctly selecting 8 out of 10 fabrics (Table~\ref{tab:tab4}), while unfine-tuned models struggled.
Selection became harder with more attributes: both failures involved three-attribute comparisons requiring finer reasoning, whereas two-attribute cases were easier. These results suggest that multi-attribute fabric selection requires structured reasoning over ranked property comparisons, with difficulty increasing as the decision involves more dimensions. Fabric-Llama-90B exhibited stepwise reasoning similar to Chain-of-Thought~\cite{Wei}. On an A800 GPU, sorting four fabrics took 2–3 minutes; future work will reduce inference time via parallelization and token optimization.

\subsection{Comparative Analysis of Baseline Methods}

\subsubsection{Symbolic Interaction Pipeline with Tactile Classification}

We implemented a tactile classification pipeline following prior work~\cite{activecloth}, where a CNN-LSTM model was trained to predict fabric attributes across three discrete levels: low, medium, and high. During deployment, the predicted attribute levels were either used in rule-based selection logic or parsed into prompts for a large language model, forming two variants: CNN IT and CNN IT*. The former employed GPT-4o for downstream reasoning, while the latter utilized LLaMA3.2-Vision-90B.
Although the tactile classifier achieved 58\% average accuracy on attribute prediction for unseen fabrics, the end-to-end scenario-level selection accuracy reached only 50\%, as shown in Table~\ref{tab:tab4}. This modular pipeline lacked access to raw sensory inputs during decision making, and could not support abstention behavior or adaptive reasoning under uncertainty. These limitations indicate the suboptimality of decoupled systems for property-sensitive material selection.

\subsubsection{Compared with Conventional Vision-Tactile MLLMs}
UniTouch~\cite{UniTouch} aligns tactile signals from multiple sensors with vision–language embedding spaces, enabling unified representations, zero-shot generalization, and descriptive generation from single-sample inputs. Its inference interface, however, is oriented toward representation alignment and description rather than explicit property-level comparison. The model does not support pairwise input, scalar ranking, or task-conditioned selection. In our evaluation, queries consistently produced generic descriptive outputs that could not be deterministically mapped to comparative or selection decisions, yielding 0\% accuracy (Table~\ref{tab:tab2}).

Octopi~\cite{Octopi} supports attribute-based prompting and can be queried in a pairwise format, but its inference is essentially categorical, with predictions made per sample and then interpreted for comparison. As shown in Table~\ref{tab:tab2}, the pretrained Octopi-13B achieved a mean accuracy of only 6\%, with meaningful predictions limited to elasticity while thickness, softness, and texture were nearly zero. When fine-tuned using either direct supervision or explanation-guided knowledge distillation, the mean accuracy improved slightly to 12.25\%, driven mainly by texture, but the prediction skewness remained extremely high (78.7). This reflects severe output imbalance and unstable comparative reasoning. Consistent with these limitations, Octopi’s downstream selection success rate reached only 40\% across ten real-world tasks (Table~\ref{tab:tab4}).

TVL~\cite{TVL} trains a touch–vision–language encoder on large-scale pseudo-labeled data, supporting open-vocabulary recognition and adjective tagging from single-sample inputs. While useful for category alignment, its outputs are free-form or tag-based rather than scalar and cannot support property-specific pairwise ranking or task-conditioned selection. Mapping these outputs into our evaluation protocol would require ad hoc thresholds and vocabulary choices, introducing non-determinism. Therefore, TVL is treated as related work but excluded from quantitative comparisons.

\subsubsection{Discussion}
These visuo-tactile systems emphasize single-sample inference, either through categorical prediction (Octopi), representation alignment (UniTouch), or open-vocabulary description (TVL), and thus lack mechanisms for property-specific pairwise scalar reasoning. Octopi, although able to accept pairwise prompts, remains limited by categorical inference, showing low accuracy, high skewness, and weak downstream selection performance. UniTouch and TVL similarly focus on descriptive or representational objectives without ranking or selection. By contrast, our framework is trained for property-level pairwise decisions under scalar attributes and supports abstention when evidence is insufficient, which improves agreement with expert judgments and the reliability of downstream selection.

\section{Conclusion}
We present \textbf{MLLM-Fabric}, a robotic framework that integrates visual, tactile, and force sensing with large language model reasoning for fabric sorting and selection. By framing material understanding as pairwise property comparisons, it enables accurate and interpretable predictions across elasticity, softness, thickness, and texture.
Tested on 220 fabrics and ten real-world scenarios, our approach outperforms symbolic, pretrained, and task-specific baselines in both property ranking and selection. This work lays the groundwork for multimodal, reasoning-enabled robotic systems.

%%%%%%%%%%%%%%%%%%%%%%%%%%%%%%%%%%%%%%%%%%%%%%%%%%%%%%%%%%%%%%%%%%%%%%%%%%%%%%%%

%%%%%%%%%%%%%%%%%%%%%%%%%%%%%%%%%%%%%%%%%%%%%%%%%%%%%%%%%%%%%%%%%%%%%%%%%%%%%%%%

\end{document}